\documentclass[sn-apa]{sn-jnl}

\usepackage{graphicx}%
\usepackage{multirow}%
\usepackage{amsmath,amssymb,amsfonts}%
\usepackage{amsthm}%
\usepackage{mathrsfs}%
\usepackage[title]{appendix}%
\usepackage{xcolor}%
\usepackage{textcomp}%
\usepackage{manyfoot}%
\usepackage{booktabs}%
\usepackage{listings}%
\usepackage{setspace}%
\usepackage[ruled,vlined,linesnumbered]{algorithm2e}
\usepackage{pdfpages}

\begin{document}

\title{FILM: Framework for Imbalanced Learning Machines based on a new unbiased performance measure and a new ensemble-based technique}
\author[1]{Antonio Guillén-Teruel}


\author[1]{\fnm{Marcos} \sur{Caracena}}

\author[1]{\fnm{Jose A.} \sur{Pardo}}

\author[1]{\fnm{Fernando} \sur{de-la-Gándara}}

\author[1]{\fnm{José} \sur{Palma}}

\author*[1,2]{\fnm{Juan A.} \sur{Botía}}\email{juanbot@um.es}

\affil[1]{\orgdiv{Departamento de Ingeniería de la Información y Las Comunicaciones}, \orgname{Universidad de Murcia}, \orgaddress{\city{Murcia}, \postcode{30100}, \state{Murcia}, \country{Spain}}}

\affil[2]{\orgdiv{Department of Neurodegenerative Disease}, \orgname{Institute of Neurology, University College London}, \orgaddress{\city{London}, \postcode{WC1N 3BG}, \country{UK}}}


\abstract{This research addresses the challenges of handling unbalanced datasets for binary classification tasks. In such scenarios, standard evaluation metrics are often biased by the disproportionate representation of the minority class. Conducting experiments across seven datasets, we uncovered inconsistencies in evaluation metrics when determining the model that outperforms others for each binary classification problem. This justifies the need for a metric that provides a more consistent and unbiased evaluation across unbalanced datasets, thereby supporting robust model selection. To mitigate this problem, we propose a novel metric, the Unbiased Integration Coefficients (UIC), which exhibits significantly reduced bias ($p < 10^{-4}$) towards the minority class compared to conventional metrics. The UIC is constructed by aggregating existing metrics while penalising those more prone to imbalance. In addition, we introduce the Identical Partitions for Imbalance Problems (IPIP) algorithm for imbalanced ML problems, an ensemble-based approach. Our experimental results show that IPIP  outperforms other baseline imbalance-aware approaches using Random Forest and Logistic Regression models in three out of seven datasets as assessed by the UIC metric, demonstrating its effectiveness in addressing imbalanced data challenges in binary classification tasks. This new framework for dealing with imbalanced datasets is materialized in the FILM (Framework for Imbalanced Learning Machines) R Package, accessible at https://github.com/antoniogt/FILM.}




\maketitle

\section{Introduction: The class imbalance problem}\label{sec:Introduction}

Machine learning is the combination of statistics, computer science and algorithms to learn patterns from data. How effective is this process is greatly influenced by  imbalanced data. In these problems, some of the concepts we aim to learn from are underrepresented in comparison with the rest. In such cases, the resulting machine learning models tend to focus on the majority class, simultaneously forgetting the minority class.

Imbalanced problems are frequent in real life  machine learning problems. This is specially true for binary classification problems \citep{21,22,23,24,25,26}. In such case, the distribution of instances between the two classes is skewed, resulting in a  disproportionate number of instances in one of the classes. All these usually leads to  biased results and poor performance. There are also imbalanced regression problems. In regression tasks \citep{27,28}, the issue of imbalanced datasets arises when the distribution of the target variable is significantly skewed. And again, the resulting machine learning model will show disproportionate better performance on the examples with better representation of the dependent variable. In definitive, it is essential to address this issue from the early stages of the data analysis tasks in any machine learning project. 

One of the domains that are critically affected by imbalances in the data is the medical domain. Examples of these are those relative to case/control cohorts of subjects of a particular  disease, intervention or monitoring activity. This is usually due to the fact that the most important class is usually the minority class. In medical datasets, the consequences of biased results and poor model performance are significantly more severe \citep{1,42}. For instance, consider the mammogram dataset example from \citep{0}, with a distribution of $99\%$ negative cases and a little $1\%$ positive cases. A standard machine learning algorithm applied to this dataset could potentially yield an impressive accuracy of $99\%$, merely by predicting all cases as negative. This approach, however, would fall short of the main goal: correctly identifying the much rarer, but crucial, positive cases while maintaining a good performance on the negative cases. These situations require of special approaches to be taken. On the one hand, it is necessary to appropriately measure performance with class-imbalance aware metrics \citep{50,52}. On the other hand, apart from a correct estimation of the performance, we also need class-imbalance aware data preparation techniques and machine learning algorithms \citep{51}.

\begin{figure}[ht!]
    \centering
    \includegraphics[width=1\linewidth]{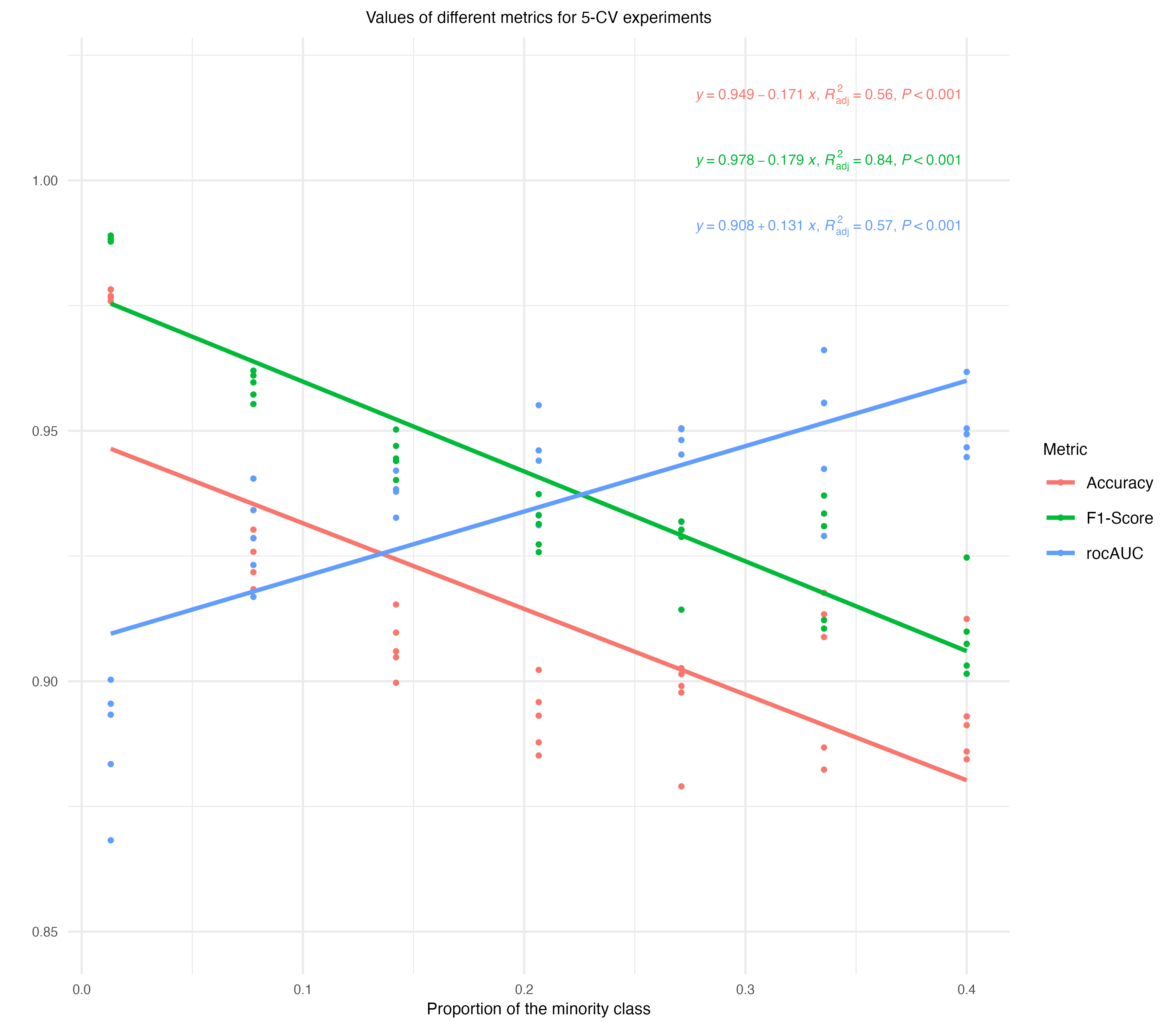}
    \caption{Values of different metrics training Random Forest with upsample for SMS dataset.}
    \label{fig:bias_inicial}
\end{figure}

One of the consequences of not using class-imbalance aware metrics is that those that are actually used become biased. Widely used  performance metrics such as accuracy, precision, recall, area under the receiver operating characteristics (rocAUC) or F1-score can be heavily biased towards the majority class in imbalanced datasets \citep{2}. We will demonstrate it through the paper but we can easily see this  Fig. \ref{fig:bias_inicial} in which we represent three of the most commont performance measures (accuracy, F1 and rocAUC) of a conventional random forest \citep{41} with automated grid-based hyperparameter optimization, used on one of the datasets we will use for experimentation (SMS, see Table \ref{table:1} and  Sec. \ref{subsec:Datasets}) in which we artificially vary the imbalance. There we see a statistically significant association between the proportion of the minority class (from now on $p_{min}$, at the x axis)  and performance values, at the y axis. The metric rocAUC even shows opposite direction with the other two. This simple experiment points to two different issues. Firstly, there is a clear association between level of imbalance and inflation (or alternatively depletion) of performance estimates. Secondly, apart from the bias, we can see certain lack of agreement across metrics as shown in previous studies \citep{43,45}. All this eventually leading to incorrect conclusions. 
Therefore, and following recommendations from previous recent work \citep{3,4} we propose using a set of complementary performance metrics through a new approach called UIC (Unbiased Integration Coefficients) whose aim is providing a single, $p_{min}$ unbiased technique, which integrates the basic metrics through a weighted sum. In subsequent sections we will empirically test the possible association of the UIC measure to $p_{min}$

Examples of widely used imbalance-aware algorithms are SMOTE, ROSE, Upsample and Downsample. You can see the previous works for a detailed description but basically, undersampling the data may  remove important examples from the dataset increasing noise whereas  upsampling tends to produce overfitting and unnecessary data overlaps. To contribute to the arena of available algorithms, we include in this work a new algorithm based on resampling the original dataset \citep{29}, which avoids the drawbacks just mentioned.  It is called IPIP (Identical Partitions for Imbalance Problems) and it creates ensembles of models trained on balanced subsets of the original. It is a new CPU-intensive approach but convenient in certain types of tasks (see Sec. \ref{subsec:IPIP}).

The rest of this paper is organized as follows. Sec. \ref{sec:Previous} surveys the diverse methodologies and methods currently being employed to address imbalance problems. Sec. \ref{sec:Materials} presents the datasets chosen for the study, the metrics chosen for the study along with the UIC algorithm, our approach with the IPIP method, the design of experiments and evaluation of the performance of the models. Sec. \ref{sec:Results} presents experimental results comparing our IPIP approach to other imbalance-aware approaches using all the proposed evaluation metrics and UIC.

\section{Previous work}
\label{sec:Previous}
Numerous strategies have been developed to tackle the inherent challenges posed by imbalanced ML problems. These strategies can be broadly categorized into resampling methods, cost-sensitive techniques, active learning and ensemble methods. Resampling artificially modify the class distribution of labeled training data to increase the balance through the production of an artificial dataset with the hope that ML algorithms will increase performance.  Undersampling techniques are aimed at balancing the class distribution by reducing the size of the majority class. This can be as simple as a random deletion of majority class examples. While it is easy to implement, it can discard potentially useful data. Oversampling techniques, on the other hand, increase the size of the minority class by adding more examples. The simplest approach is to randomly replicate minority class instances, which can lead to overfitting as it simply replicates the same data \citep{2,10}.

An slightly more complex resampling method is SMOTE (Synthetic Minority Over-sampling Technique) \citep{11}. It works by selecting examples that are close in the feature space, drawing a line between the examples in the feature space and creating a new sample at a point along that line. This way, the minority class is over-sampled by introducing synthetic examples along the line segments joining any or all k minority class nearest neighbors. While SMOTE is very effective, it has limitations including the introduction of artificial noise and the oversampling of borderline examples. A SMOTE extension called Adaptative Syntetis Sampling (ADASYN) \citep{12} is a technique that generates minority data samples based on their distributions. It focuses on generating synthetic data for difficult examples, which are minority class examples closer to majority examples. This approach helps shift the decision boundary of the classifier towards challenging examples and improves the learning algorithm’s performance on imbalanced datasets. ADASYN shares the same idea as SMOTE, but it focuses on generating a higher number of synthetic examples around minority class examples closer to the majority class. These examples contain, intuitively, more interesting and non-redundant information. An usual alternative to SMOTE is ROSE (Random Over-Sampling Examples) \citep{18}, which operates by generating artificially balanced samples using a smoothed bootstrap technique (i.e., random sampling with replacement of cases in the minority class).

Although the algorithms mentioned above are the most widely used in the literature and are the ones that we compared with the new IPIP model, there are other types of methods for dealing with a class imbalance that help us to give context to IPIP and see what type of model it is. Cost-sensitive learning applies an integral approach where misclassification costs are embedded in the learning process, particularly focusing on the correct classification of the minority class. These costs may be calculated statistically or given by a domain expert in the form of a cost matrix. MetaCost \citep{13} incorporates the cost matrix into the learning process itself treating all classifiers as black boxes, therefore it is applicably virtually to any ML algorithm. The Cost-sensitive Weighted approach \citep{14} is a method that assigns each instance a weight proportional to its misclassification cost. This is usually manually defined in a cost matrix. The weights bias the classifier towards correctly classifying instances with higher misclassification costs by a combination of upsampling high-cost instances and downsampling low-cost ones. MVCSKL\citep{49} uses an asymmetric linear-exponential loss function in a cost-sensitive kernel learning framework to assign diverse weights to distinct samples for different classes in a dataset with incomplete information. Active learning is a learning paradigm that allows the model to interactively acquire labels for uncertain data points from an oracle, like a human annotator or a side selection algorithm. Within imbalanced problems, active learning seeks to acquire labels for the most informative instances, typically those near the decision boundary between classes. For example, margin-based active learning with Support Vector Machines (SVM) \citep{15} aims to identify and learn from instances which are closest to the hyperplane of separation as their classification is more uncertain for this algorithm. Fair Active Learning using fair Clustering, Uncertainty, and Representativeness (FAL-CUR)\citep{46} that address fairness concerns in Active Learning while maintaining model accuracy by combining uncertainty and representative sampling with fair clustering. Online learning algorithms (OLA) continuously process new data, instance-by-instance and integrate it without retraining on all previous data. OLA approaches can be used within imbalanced problems \citep{16} through ISS (Instance Selection Strategies). ISSs are used to decide on the new points to be added to the model in the active learning process. CSRDA\citep{47} is a cost-sensitive and an online algorithm that achieve a balance between low classification cost and high sparsity, which was applied to three online anomaly detection tasks. Ensemble techniques \citep{33} are based on the integration of basic machine learning models into meta-models. Ensembles are particularly effective in dealing with imbalance through bagging and boosting styles of ensemble construction. Bagging \citep{32} constructs the basic models from resamples of the original training data with repeated samples and samples left out as a result. Specifically,  UnderBagging combines the principles of bagging and under-sampling. It creates a number of balanced subsets, or bags, by undersampling the majority class. From each bag, a ML model is created. Final predictions are made by combining in some way the individual predictions from the components of the meta-model. Variants are Roughly Balanced Bagging \citep{31}, based on creating resamples based on the  inverse binomial distribution, and OverBagging, i.e., bagging with oversampling. SMOTEBagging \citep{34} combines bagging resampling with  SMOTE to guarantee their balance.  While in bagging the basic models are build in parallel, in boosting they are created sequentially instead as the current model focuses on modeling the errors from the previous model.  This process continues untils there is practically no error to model. For example, SMOTEBoost \citep{30} combines boosting with SMOTE. We can also find bagging-boosting hybrids like, for example, EasyEnsemble \citep{17}. It is a bagging algorithm that uses Adaboost (a boosting approach) as the basic learner, which implies that each boosting process can be performed in parallel. Another hybrid algorithm is BalanceCascade \citep{17} and requires sequential model creation. This time, within each iteration, instances of the majority class that are classified with high confidence are purged from the dataset. This leads to a progressive reduction in the size of the majority class, helping to prevent overfitting and sharpening the model’s focus on instances that are more challenging to classify correctly. HICD \citep{48} is another hybrid algorithm that uses data density to perform a sampling of the data, generates subproblems for different classes of instances to construct ensemble models, and selects appropriate classification models based on the instance distribution.

\begin{figure}[!ht]
    \centering
    \includegraphics[width=1\linewidth]{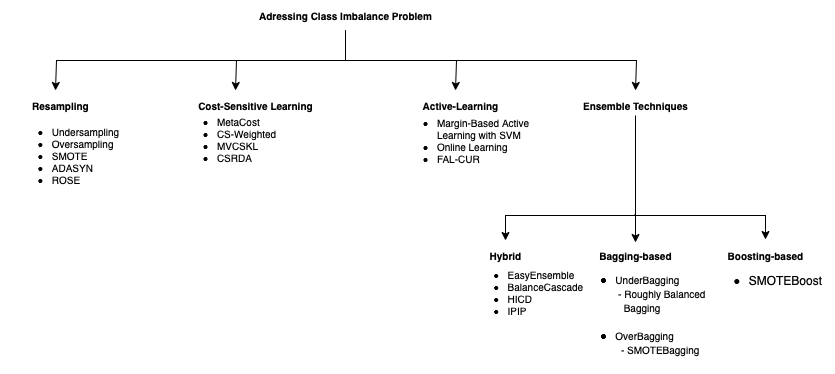}
    \caption{Taxonomy of the imbalance learning techniques studied in this section.}
    \label{fig:Taxonomy}
\end{figure}

A full taxonomy of the imbalance learning techniques mentioned in this section can be seen in Fig. \ref{fig:Taxonomy}. The algorithm we propose in this work, IPIP, is an hybrid ensemble-based algorithm.

\section{Materials and methods}
\label{sec:Materials}
\subsection{Datasets}
\label{subsec:Datasets}
We performed experiments on seven different datasets, with a variety of class imbalance ratios and sample and feature sizes. This heterogeneity improves the assessment of robustness of our approach. Table \ref{table:1} shows a summary of each dataset ordered by Imbalanced Ratio (IR), which measures the extent of imbalance in a dataset. IR is defined as the fraction of the sample size of the majority class over the sample size of the minority class.

All datasets are accessible from external sources \citep{6,7,8,35,36,37} except the SMS dataset, which was built in-house and was provided by The Regional Health System (SMS) from the Region of Murcia (Spain). It consists on 2 demographic features and 19 comorbidities obtained from 86,867 electronic medical records of COVID-19 patients and it is for binary classification, i.e., whether a patient survives or not. The number of positive cases is 1141. Please note that the Forest dataset \citep{35} is a multiclass dataset with seven classes. We only maintened the two larger classes to make it binary. Satimage  had six classes. We selected the minority class, and a synthetic one made of all other samples.

\begin{table}[!ht]
\resizebox{\textwidth}{!}
{
\begin{tabular}{|p{2cm}|p{2cm}|p{1cm}|p{2cm}|p{2cm}|p{2cm}|p{2cm}|p{2cm}|}
\hline
\multicolumn{1}{|c|}{\textbf{Datasets}} & \multicolumn{1}{c|}{\textbf{Instances}} & \multicolumn{1}{c|}{\textbf{IR}} & \multicolumn{1}{c|}{\textbf{Attributes}} & \multicolumn{1}{c|}{\textbf{Categorical atts.}} & \multicolumn{1}{c|}{\textbf{Min labels}} & \multicolumn{1}{c|}{\textbf{Max labels}} & \multicolumn{1}{c|}{\textbf{Mean labels}} \\
\multicolumn{1}{|c|}{} & \multicolumn{1}{c|}{} & \multicolumn{1}{c|}{} & \multicolumn{1}{c|}{} & \multicolumn{1}{c|}{} & \multicolumn{1}{c|}{\textbf{in cat. att.}} & \multicolumn{1}{c|}{\textbf{in cat. att.}} & \multicolumn{1}{c|}{\textbf{in cat. att.}} \\ \hline
SMS                                    & 86867                                & 75.34                                & 22                                         & 19                                                   & 2                                        & 2                                        & 2                                         \\ \hline
Mammograhy                             & 11183                                & 42.10                                & 7                                          & 0                                                    & 0                                        & 0                                        & 0                                         \\ \hline
Forest                                 & 38501                                & 13.03                                & 28                                         & 17                                                   & 2                                        & 2                                        & 2                                         \\ \hline
Satimage                               & 6430                                 & 9.29                                & 37                                         & 0                                                    & 0                                        & 0                                        & 0                                         \\ \hline
Adult                                  & 48842                                & 3.18                                & 15                                         & 8                                                    & 2                                        & 42                                       & 12.75                                     \\ \hline
Phoneme                                & 5404                                 & 2.41                                & 6                                          & 0                                                    & 0                                        & 0                                        & 0                                         \\ \hline
Diabetes                               & 768                                  & 1.87                                & 9                                          & 1                                                    & 17                                        & 17                                       & 17                                       \\ \hline
\end{tabular}
}
\caption{Datasets used in the paper, including information for the number of instances, the imbalanced ratio (IR), number of attributes, number of categorical attributes and details on label variety for those attributes.}
\label{table:1}
\end{table}

\subsection{Performance metrics}
\label{subsec:Metrics}
We have multiple ways we can use to inspect binary classifiers performance but all are based on four elements, depending on whether the  outcome from the classifier is correct: true positives (TP) and true negatives (TN) refer to positive and negative instances, correctly classified, respectively.  On the other hand, false positives (FP) and false negatives (FN) are analogous for incorrectly classified instances. We can compound a confusion matrix with these four values for a fine grain judgement on classifier performance.  Moreover, we can combine those to obtain a variety of performance metrics (see Table \ref{table:equations}). Accuracy is defined as the proportion of correctly classified instances of the positive and negative classes. Precision is the fraction of all relevant instances divided by the instances obtained. Recall is the fraction of relevant instances obtained over the total number of relevant instances. The F1-score is obtained through the average of two complementary measures: accuracy and recall. Cohen's kappa\citep{19} is a measure for evaluating the accuracy of a classifier that tries to take into account the imbalance between positive and negative examples. The Matthew's correlation coefficient \citep{20} measures the difference between the predicted values and actual values and is equivalent to a $\chi^2$-square statistics for a $2\times 2$ confusion matrix. The sensitivity is the proportion of correctly classified positive instances and the specificity is the proportion of correctly classified negative instances . Balanced Accuracy is calculated as the average of sensitivity and specificity. The Geometric mean  is calculated as the geometric mean of the sensitivity and specificity. Two well-known metrics are the Area Under the Receiver Operating Characteristic Curve (rocAUC) and the Area Under the Precision-Recall curve (prAUC), where the Receiver Operating Characteristic Curve is a representation of the true positive rate (sensitivity) as a function of the false positive rate for different probability values. Precision-Recall curve shows the precision as the function of recall. Although the Imbalanced Ratio (IR) is the most primarily used metric for measuring dataset imbalance, we will mainly use the proportion of the minority class, $p_{min}$, going forward. This is because when we work with it, we have the restriction that IR is not unbounded, while $p_{min} \in (0,0.5)$ and $p_{min}$ is easier to interpret. The relationship between IR and $p_{min}$ in a dataset can be expressed as follows: 

\begin{equation}
\label{eq:IR-pmin}
p_{min}=\frac{1}{IR+1}.
\end{equation}

Of all the metrics mentioned above, we will work with Accuracy, Cohen's Kappa, Balanced Accuracy, F1 Score, rocAUC, prAUC, Mathew's Correlation Coefficient and Geometric Mean. We will see all these metrics show variable biases towards $p_{min}$.

\begin{table}[!ht]
\begin{spacing}{1.3}
\resizebox{\textwidth}{!}
{
\begin{tabular}{|l|c|}
\hline
\multicolumn{1}{|c|}{\textbf{Metric}} & \textbf{Equation}                                                                            \\ \hline
Accuracy (Acc)                             & $ \frac{TP+TN}{TP+FP+TN+FN}$                                                            \\
 Sensitivity&$ \frac{TP}{TP+FN}$\\
 Specificity&$ \frac{TN}{TN+FP}$\\\hline
Precision (precision)                            & $\frac{TP}{TP+FP}$                                                                 \\ \hline
Recall ($recall$)                               & $\frac{TP}{TP+FN}$                                                                    \\
 False positive rate (FPR) &$ \frac{FP}{FP+TN}$\\ \hline
F1-Score ($F_1$)                             & $2\cdot \frac{precision\cdot recall}{precision+recall}$                                  \\ \hline
Cohen's Kappa (Kappa)                        & $\frac{2\cdot (TP\cdot TN - FN \cdot FP)}{(TP+FP)\cdot (FP+TN)+(TP+FN)\cdot (FN+TN)}$ \\ \hline
Mathew's correlation coefficient (MCC)     & $\frac{RN\cdot TP- FN \cdot FP}{\sqrt{(TP+FP)\cdot (TP+FN)\cdot (TN+FP)\cdot (TN+FN)}}$  \\ \hline
Balanced Accuracy ($Bal\_Acc$)                    & $\frac{sensitivity + specificity}{2}$                                             \\ \hline
Geometric mean (GEOM)                       & $\sqrt{sensitivity \cdot specificity}$                                                  \\ \hline
\end{tabular}
}
\end{spacing}
\caption{Equations of evaluation metrics for binary classification.}
\label{table:equations}
\end{table}

\subsection{The Unbiased Integration Coefficients (UIC) method}
\label{subsec:UIC}
In most of data analysis problems that involve ML, there is a step in the analysis pipeline devoted to algorithm´s hyperparameter optimization and model selection. On the one hand, we need to find the best values for the algorithm parameters given the target dataset. For such purpose, we usually create a set of candidate models by exploring the hyperparameter space \citep{38,39} to select those hyperparameter values used at the best model created in that exploration. On the other hand, we have to trust the final model we will use in exploitation mode. In both of these processes, it is crucial to use an appropriate statistic to measure performance. This issue is even more crucial when classes are imbalanced because, as we have seen, the possible statistics may be biased towards the proportion of the minority class, $p_{min}$. Our main thesis is that we can create a new measure, as an integration of all the biased ones, through an additive model in which all basic measures appear inversely weighted by their respective correlation with $p_{min}$ with the intent that the new statistic will be less biased to $p_{min}$. To do this, we need an estimate of this correlation. To obtain this estimate, we proceed as follows. Let's say we have $k$ performance metrics, for example, all those which appear at table \ref{table:equations}. We want to create a joint metric, as uncorrelated as possible with $p_{min}$. Let us denote with $p_d$ the original minority class proportion in a specific dataset $d$. We then create $n$ datasets, where $n$ is an even number and $n\geq 6$, with artificial class proportions by randomly sampling the entire original dataset. If $p_d > 0.05$, then $n/2$ of those datasets have variable, and equally distributed, class proportions within the interval $[0.05, p_d]$ and the other $n/2$ datasets are the same within the interval $[p_d,0.4]$. Otherwise, if $p_d \leq 0.05$, $n$ datasets are created in the same way within the interval $[p_d,0.4]$. $p_d$ values greater than 0.4 are not considered proper of imbalanced datasets and, therefore, it would makes no sense to use UIC with them.

Given an original dataset $d$ and its $n$ sampled versions, we split each of them into training and test sets and then train ML models on all of them, obtaining $n+1$ models. We evaluate each model in its respective test set using the $k$ different metrics. Let us denote the vector of all the metrics for a given dataset $d$ with $\mathbf{m_{dk}}= (m_{dk},m_{1k},m_{2k},...,m_{nk})$ and the vector of of minority class proportions with $\mathbf{p_{min}} = (p_d,p_1,p_2,...,p_n)$. To obtain an estimate of the bias of the $k$-th metric in a particular dataset $d$, we can calculate the Pearson's Correlation, $r_{dk}$, between $\mathbf{m_k}$ and $\mathbf{p_{min}}$.

Thus, each metric is assigned a weight according to its correlation with $\mathbf{p_{min}}$, so that the closer $r_{dk}$ is to 0, the higher its weight should be. Therefore, to assign a weight to a given metric, we need a function that, when $r_{dk}$ is close to $0$ returns a weight close to $1$, and, otherwise, when $r_{dk}$ is close to $\pm 1$, returns a weight close to $0$. Such a function can be modelled through a Gaussian, i.e., an exponential that takes values from a concave quadratic function with parameters $(a,b,c)$ (Eq. \ref{eq:gaussian}), being $a$ the curve max height, $b$ its centre where the function reach its maximum value, and $c$ its width:

\begin{equation}
\label{eq:gaussian}
  G(x)=\displaystyle a \cdot \exp\left(-\frac{(x-b)^2}{2c^2}\right)  
\end{equation}

\begin{figure}[!ht]
    \centering
    \includegraphics[width=1\linewidth]{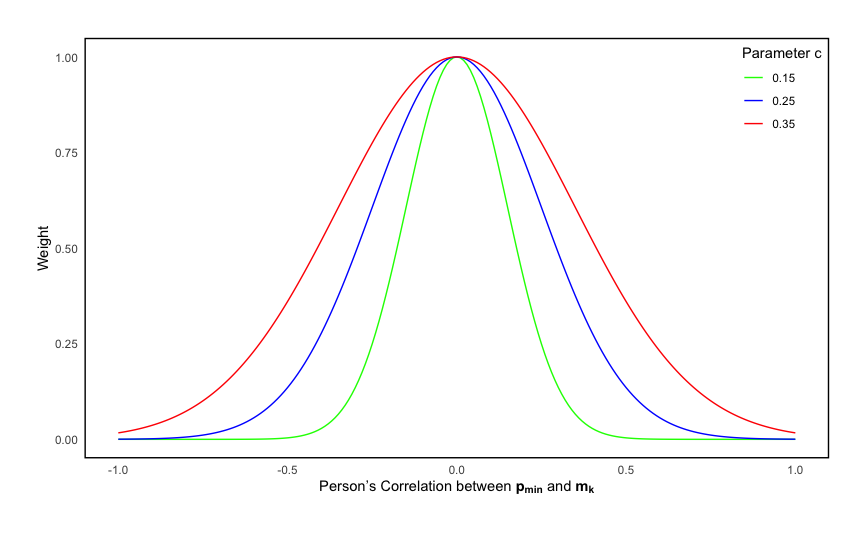}
    \caption{Three different Gaussian functions. Red curve with parameters $(1,0,0.35)$, blue curve with parameters $(1,0,0.25)$ and green curve with parameters $(1,0,0.15)$.}
    \label{fig:Gaussiana}
\end{figure}

In Fig. \ref{fig:Gaussiana}, three different Gaussian functions are shown with different values for $c$, and with a maximum value $a=1$ and centred in 0, $c=0$. Pearson correlations values of $x$ close to $0$, lead to metric weights close to $1$. Extreme Pearson values for $x$, i.e., $\pm 1$ lead to weights close to $0$. Note that values of $c$ close to $0$ lead to a stronger penalty for $p_{min}$ bias near absolute values of 1. For example, in the figure, the curve for $c=0.15$ generate weight values closer to 0 than those we get with $c=0.35$ (see results). In Table \ref{table:2}, there are several values of UIC that are $0$, which means the metric is strongly biased towards $p_{min}$.

Therefore, given an imbalanced dataset $d$ and a set of correlations, $\mathbf{r_d}=\{r_{d1},\ldots,r_{dk}\}$, between the vector measures $\mathbf{m_{dk}}$ and the vector of minority class proportions $\mathbf{p_{min}}$, and a set of weights for each measure obtained by applying the function $G(x)$ to the set of correlations, that is, $\mathbf{w_d}=\{w_1=G(r_{d1}),\ldots,w_k=G(r_{dk})\}$, the UIC metric can be defined as the weighted sum of each metric (Eq. \ref{eq:UIC}).

\begin{equation}
    UIC=\sum_{j=1}^{k}w_j\cdot m_{dj}.
\label{eq:UIC}
\end{equation}

\begin{figure}[!ht]
    \centering
    \includegraphics[width=1\linewidth]{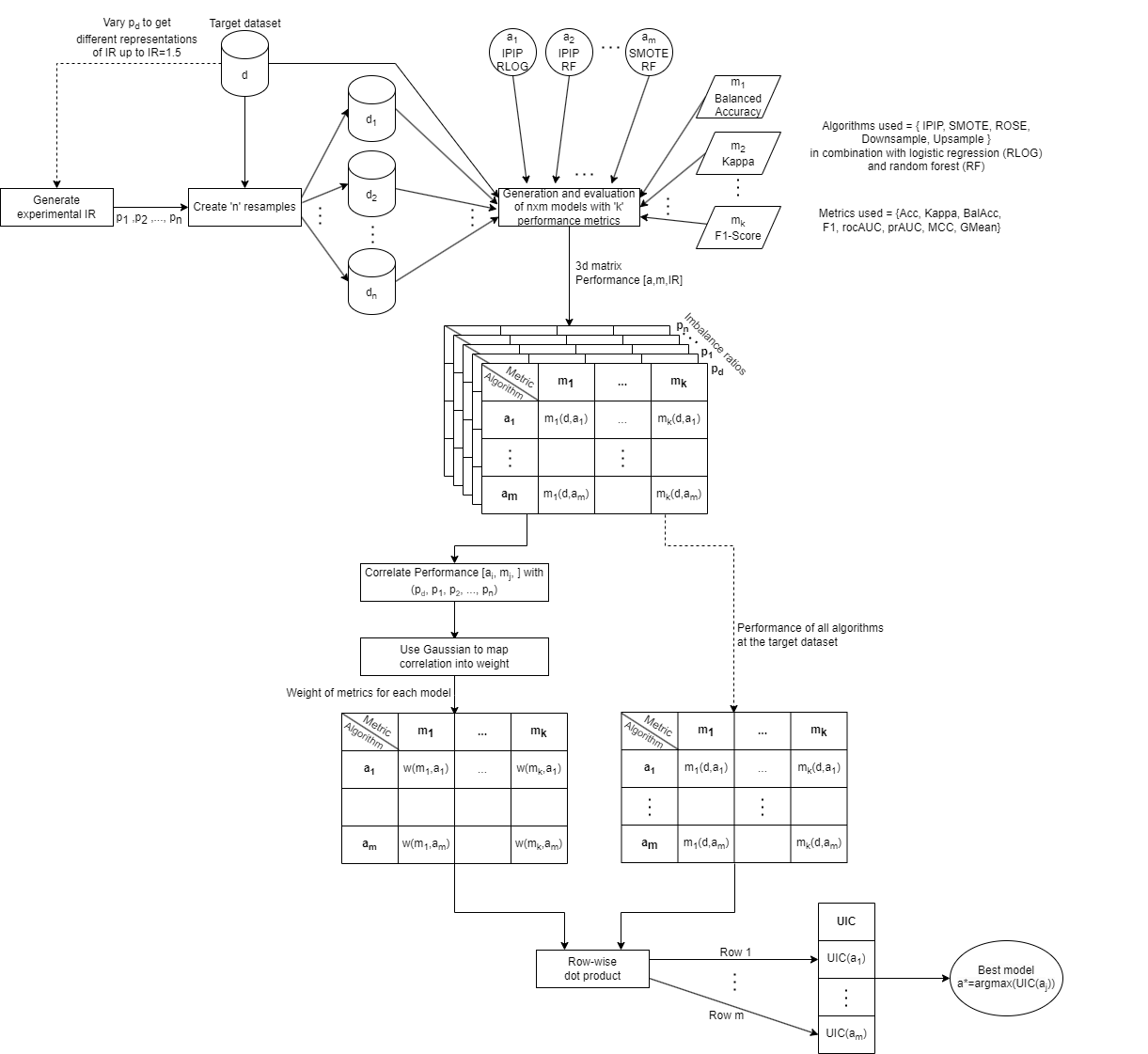}
    \caption{FILM methodology. First, $n$ datasets are obtained from the original dataset $d$ through sampling with several imbalance ratios. A set of algorithms is then trained on $d$ and its resamples, resulting in $k$ metrics for each algorithm on each dataset. Thus, a three-dimensional matrix is formed to retain this values and then we aggregate the metric results for each algorithm across all $n+1$ datasets and retain the $k$ metric values of all algorithms in $d$. This provides weights for each metric and algorithm. The weights and values of the metrics are finally aggregated over the original dataset to obtain the UIC metric. This allows us to determine which algorithm has obtained the maximum value.}
    \label{fig:FILM}
\end{figure}

This metric can be obtained for several ML algorithms and choose a better algorithm for our problem based on UIC. This methodology can be seen in Fig. \ref{fig:FILM} and can be carried out using the FILM (Framework for Imbalanced Learning Machines) R Package, accessible at \url{https://github.com/antoniogt/FILM}.

\subsection{The IPIP algorithm}
\label{subsec:IPIP}

Ensemble methods offer a valuable solution for complex binary classification problems by enabling multiple models to be trained from resampled training data. When constructing an ensemble for tackling imbalanced problems, it is advisable to (1) avoid generating synthetic samples and (2) employ base learners as black boxes, thereby facilitating the use of any base learner on balanced resamples. To address the issue of imbalanced data, we present a novel technique termed IPIP (Identical Partitions for Imbalance Problems). IPIP generates multiple resamples by down-sampling the majority class with replacement, resulting in a balanced data set. Subsequently, each balanced resample has the potential to generate a new model, which is integrated into an ensemble through majority voting. The algorithm ensures that all examples from the minority class are selected at least once. IPIP's task is to balance the dataset and build an ensemble classifier.

The algorithm for IPIP is outlined in Algorithm \ref{pseudo:IPIP}. Firstly, the initial dataset is divided into training and test sets, each containing the same class proportion as the original dataset being the training set composed of the $p_{holdout}$ proportion of the total number of instances (Algorithm \ref{pseudo:IPIP} line 2). To ensure representation of all samples from the minority class throughout the resamples, IPIP divides the training data into $b_s$ balanced subsets, with $p_{min} \approx 0.45$ (see Algorithm \ref{pseudo:IPIP} line 3). Each of these subsets contains non repeated and non synthetic samples from the original dataset and is used to develop one or more models. These models are then combined into a majority voting ensemble for classification.

IPIP ensures, with a confidence level of $(1-\alpha)\cdot 100\%$, that every single sample in the minority class is present in at least in one of the $b_s$ balanced subsets during data preparation. To achieve this, the value for $b_s$ ought to be calculated in the following manner. Let $n_{min}$ represent the minimum number of samples from the minority class required for the $b_s$ resamples. We set the number of samples in the majority class, $n_{max}$, so that $n_{max}=n_{min}+n'$. Here, $n'$ denotes the quantity of instances exceeding $n_{min}$ that we need from the majority class to attain a value of $p_{min}\approx 0.55$ in every resample. Bear in mind that as $n_{min}$ gets closer in each resample to $n_{min}$ in the original dataset, $b_s$ decreases. During our experiments, we established $n_{min}$ to reflect 75\% of the minority class samples in the original dataset. More specifically, suppose that $n$ denotes the number of samples in the minority class of the original dataset. To obtain a set of $n_{min}$ samples, where $n_{min}<n$, from the original minority class, we randomly select samples with replacement uniformly so that, in each iteration, each sample has a probability of $1/n$ of being chosen. In a replacement sampling, iterations are independent, therefore the random variable $X_k$ indicating the number of times any given sample is selected for the $k$-th resample, with $k\in\{1,...,b_s\}$, follows a binomial distribution with parameters $(n_{min},1/n)$ (Eq. \ref{eq:binomial})

\begin{equation}
    P\left [ X_k = r \right ] = \binom{n_{min}}{r}n^r (1-1/n)^{n_{min}-r}
    \label{eq:binomial}
\end{equation}

Since we want that each sample in the minority class should be selected at least once, we have:

\begin{equation}
    P\left [ X_k > 0 \right ] = 1-(1-1/n)^{n_{min}}=q
\end{equation}

We can define another random variable, $Y_k$, as follows:

\begin{equation}
Y_k = \begin{cases}
    1 & \text{if the sample is chosen for the k-th subset}\\
    0 & \text{otherwise}
\end{cases}
\end{equation}

In this scenario, $Y_k$ follows a Bernoulli distribution with parameter $q$. As each $Y_k$ is independent, the total number of subsets in which a specific sample appears, $Y$, can be computed as $Y = Y_1 +...+ Y_{b_s}$. This follows a binomial distribution of parameters $(b_s,q)$. Hence, the probability of an instance being chosen to be included in one of the $b_s$ subsets is:

\begin{equation}
    P\left [ Y>0 \right ]=1-(1-q)^{b_s}
\end{equation}

We want this probability to be greater than $\alpha$, therefore $b_s$ must satisfy:

\begin{equation}
\label{eq:bs}
    b_s > \frac{log(1-\alpha)}{n_{min}\cdot log(1-1/n)}
\end{equation}

Now, let define a function $f^n_\alpha:N \rightarrow R$ such that 
\begin{equation}
f^n_\alpha(n_{min})=\frac{log(1-\alpha)}{n_{min}\cdot log(1-1/n)}
\end{equation}

This function is lower bounded by the number of balanced subsets $b_s$ and determines the number of balanced subsets required to satisfy that, with probability greater than $\alpha$, every sample of the minority class appears, at least once, in one of the $b_s$ subsamples.

For each balanced subset, the algorithm builds an ensemble that progressively accumulates up to $b_E$ ML models (algorithm \ref{pseudo:IPIP} line 8). To do this, we have defined a maximum number of possible attempts to expand the ensemble. This number is given by the function $m_t(|T|)$, which depends on the number of base models $T$ already ensembled for a given resampling and must be a monotonically decreasing function (algorithm \ref{pseudo:IPIP} line 8). In our experiments we used the function described in Eq. \ref{eq:mt}, but any discrete decreasing function of $|T|$ could be used.

\begin{equation}
\label{eq:mt}
    m_t(|T|) = \left \lceil  \frac{b_E-|T|}{3}\right \rceil
\end{equation}

IPIP creates diverse resamples to enable models to target different subproblems (Algorithm \ref{pseudo:IPIP}, line 9). To this end, IPIP subsamples with a $p_{min} \approx0.5$ in each resample and $n_{min}$ instances from the minority class. And as previously demonstrated, we only need a maximum of $b_E$ models. Similar to the process of obtaining the lower bound for $b_s$ (as shown in Eq. \ref{eq:bs}), $b_E$ must also fulfill Eq. \ref{eq:bE}.
 
\begin{equation}
b_E > \frac{log(1-\alpha)}{n_{min}\cdot log(1-1/n_{min})}.
\label{eq:bE}
\end{equation}

New base models (line 13 of Algorithm \ref{pseudo:IPIP}) are incorporated into the ensemble only if they enhance the overall performance on the test set (line 12 of Algorithm \ref{pseudo:IPIP}) that has a class distribution similar to the original dataset. The voting scheme assigns an instance to the majority class if a minimum of 75\% of the models in the ensemble classify it as such. Otherwise, they are classified as belonging to the minority class (algorithm \ref{pseudo:IPIP} line 11). The final model is formed by combining the ensembles created from each balanced subset (line 20 of Algorithm \ref{pseudo:IPIP}).

Finally, new instances are classified by individually evaluating each of the ensembles that have been built for each balanced subset. The final model classifies a sample as negative if $50\%$ of the ensembles classify it as negative (majority class).

\begin{algorithm}[H]
\caption{IPIP($X,p_{min},b_s,b_e,ML,tries_{max}$,$p_{hold\_out}$)}
\label{pseudo:IPIP}
\SetAlgoLined
\KwIn{$p_{min}$}
\KwIn{$b_s$: number of subsamples of the original set with minimum value given by Eq. \ref{eq:bs}}
\KwIn{$b_E$: maximum number of models in an ensemble}
\KwIn{$ML$: ML technique to be used}
\KwIn{$tries_{max}$: function that sets the maximum number of attempts at ensemble construction}
\KwIn{$p_{hold\_out}$}
\KwOut{$M$: Ensemble of $ML$ models}
$M \gets \varnothing$\;
$D^{train},D^{test}$$ \gets stratified\_hold\_out(X,p_{hold\_out})$\;
$\{D^{train}_1,D^{train}_2,...,D^{train}_{b_s}\} \gets balance\_subsampling(D^{train},p_{min})$\;
\ForEach{$D \in \{D^{train}_1,D^{train}_2,...,D^{train}_{b_s}\}$}{
  $M_{bs} \gets \varnothing$\;
  $tries \gets 0$\; 
  $best\_performance \gets 0$\;
  \While{$|M_{bs}|< b_E$ \textbf{and} $tries < tries_{max}(|M_{bs}|)$}{
    $D' \gets balance\_subsampling(D,0.5)$\;
    $model \gets train(D',ML)$\;
    $performance \gets evaluate(M_{bs} \cup \{model\},D^{test})$\;
    \eIf{$performance > best\_performance$}{
        $M_{bs} \gets M_{bs} \cup \{model\}$\;
        $best\_performance \gets performance$\;
        $tries \gets 0$\;
    }{
        $tries \gets tries+1$\;
    }
  }
  $M \gets M \cup \{M_{bs}\}$\;
}
\KwRet{M}\;
\end{algorithm}

\subsection{Experiments}
\label{subsec:Experiment}

Seven binary classification datasets were used in the experiments (see table \ref{table:1}). The simplest non-overfitting logistic regression and random forests were selected as base learners in all experiments implemented. The model-building process has been implemented using R package \textit{caret} \citep{40} (version 6.0-94), and \textit{glm} for logistic regression (R version 4.3.1) and \textit{ranger} for Random Forest (version 0.15.1). Two distinctly different modelling approaches were tested. To evaluate the effectiveness of IPIP, we compared it against well-known Imbalance-Aware Approaches (IAA) such as Smote, Rose, Upsample, and Downsample. Additionally, we incorporated two ensemble-based hybrid techniques, SmoteBoost and UngerBagging, which share similarities with IPIP's approach. All experimentation utilized a five-fold cross-validation strategy. For the Random Forest model, a hyperparameter search strategy has been conducted for each dataset. During the hyperparameter search, Cohen's Kappa was used as a performance metric as it presents a better trade-off between simplicity and $p_{min}$. The hyperparameters used for each dataset are described in Table 1 in the supplementary material.

\subsubsection{Agreement and disagreement plots between metrics: Concordance plots}
\label{subsec:Plots}

The first question that needs to be addressed is to determine whether all of these metrics that have been evaluated in our experiments are consistent in identifying a particular IAA technique as the best one for a given data set. This was accomplished by applying all IAA algorithms to all datasets, building logistic regression and random forest models and collecting all metrics using a 5-fold cross-validation procedure, varying the proportion of the minority class (see Sec. \ref{subsec:UIC}). To ensure results can be analysed in a visually interpretable manner, we propose a new type of plot called concordance plot, which represents the proportion of agreement between metrics for a given IAA technique and model pair on a given dataset.

Given a dataset, $d$, we apply a set of IAA techniques $A=\{a_1,a_2,...,a_n\}$ and evaluate them with a set of performance metrics $M=\{m_1,m_2,...,m_m\}$ obtained through a set of experiments $E=\{e_1,e_2,....,e_t\}$. Each experiment consists of building a specific machine learning model for each of the different data sets sampled from $d$, with different minority class proportions, $\mathbf{p_{min}}$, using 5-fold cross-validation resampling. In this context, we can establish that there is an agreement or concordance between the metrics $m_i\mbox{, }m_j\mbox{, }i\neq j\mbox{, }1\leq i\mbox{, }j\leq m$ for a specific experiment $e \in E$ if both metrics give the maximum score to the same technique $a_k \in A$ on $e$. Otherwise, we consider that there is a disagreement or discordance. To construct a concordance plot, we need to compute the win ratios for each technique $a_k$ using a particular metric $m_i$ across all experiments $e_t \in E$. This can be achieved by calculating the number of times $a_k$ technique score better for $m_i$ metric in each experiment $e$ and dividing it by the total number of experiments, $|E|=|\mathbf{p_{min}}|\times 5$. Agreement ratios can be visually represented via pie charts for each metric pair $m_i,m_j$, where each slice corresponds to technique $a_k\in A$. A fully coloured pie denotes agreement between $m_i$ and $m_j$, whereas an empty or nearly empty pie indicates strong disagreement. Conversely, the larger slice indicates the best performing algorithm on the data set $d$ for the cases where $m_i$ and $m_j$ agreed. To produce a concordance plot (Fig. \ref{fig:fig1}), we collect all the slices in an upper triangular matrix, where each $(i,j)$ index, $1\leq i,j\leq m$, shows the agreement/disagreement between the metrics $m_i$ and $m_j$. The main diagonal represents complete agreement as it compares a metric with itself.

\subsubsection{Correlation between metrics and proportion of the minoiry class \texorpdfstring{$p_{\text{min}}$}{p\_min}}
\label{subsec:Experiments2_Correlations} 

A key question in this paper is whether the most commonly used metrics for evaluating binary classification models are biased by the proportion of the minority class $p_{min}$ in a given dataset $d$ and ML model. To shed light on this fundamental question, we investigate the correlation between the set of metrics $\mathbf{m_{dk}}$ acquired by evaluating the machine learning model utilizing a specific IAA method and the set of variations in minority class proportions, $\mathbf{p_{min}}$, employed to produce diverse versions of the initial dataset $d$. This procedure is explained in more detail in Sec. \ref{subsec:UIC}.

\subsubsection{Comparison of metrics biases towards the proportion of minority class \texorpdfstring{$p_{\text{min}}$}{p\_min}}

\label{subsec:Experiments3_UIC_Cor}

In this paper, we introduce a new performance metric called UIC (see Sec. \ref{eq:UIC}), which exhibits less bias towards minority class proportions compared to other metrics employed. To empirically verify this, we incorporate the correlation between UIC and $\mathbf{p_{min}}$ in the correlation vector $\mathbf{r_{dk}}$, for each dataset $d$. If the resulting vector $\mathbf{r_{dk}}$ equals the zero vector, it implies that the $k$-th metric is unbiased towards $p_{min}$, at least for the dataset $d$. Therefore, a measure of non-bias can be determined by calculating the Euclidean distance of vector $\mathbf{r_{dk}}=<r_{d_0i},r_{d_1i},\ldots, r_{d_ni}>$ from the zero vector.  To compare the bias of UIC with the other metrics, we visually evaluate the distribution of values for the Pearson's correlation value between metrics and $p_{min}$ in relation to the various datasets using boxplots. Additionally, pairwise statistically significant differences are assessed between these correlation values of UIC and the other metrics through a non-parametric pairwise Wilcoxon Hypothesis Test, with Bonferroni correction applied.

\section{Results}
\label{sec:Results}
\subsection{No agreement among metrics to choose the best model}
\label{subsec:Q1}

\begin{figure}[ht!]
    \centering
    \includegraphics[width=0.8\linewidth]{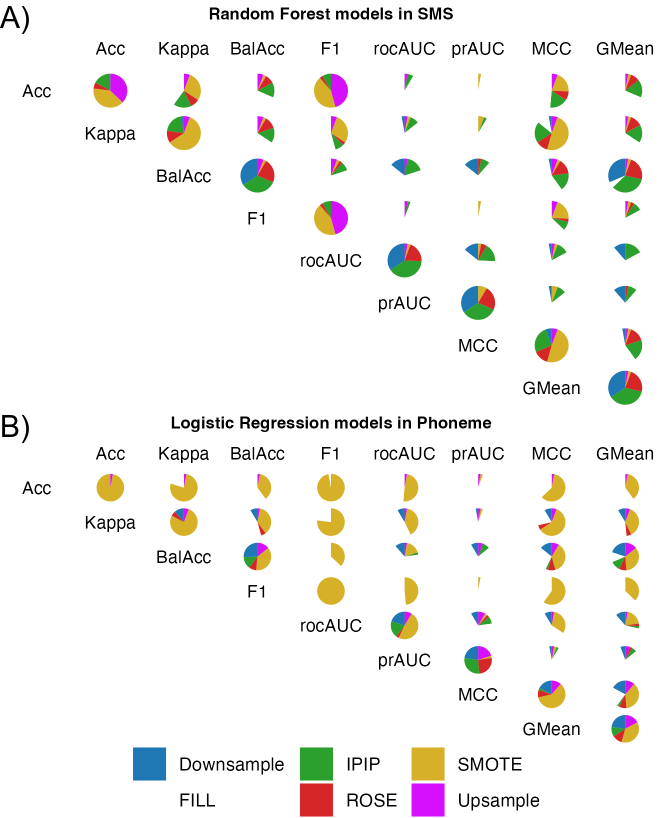}
    \caption{A) Concordance plot of the SMS dataset using Random Forest. B) Concordance plot of the Phoneme dataset using Logistic Regression models.}
    \label{fig:fig1}
\end{figure}

We have produced agreement plots for the two basic learning methods, namely logistic regression and random forest, for all 7 of the datasets examined (see Sec. \ref{subsec:Datasets}). For instance, Fig. \ref{fig:fig1} displays the pairwise agreements (see Sec. \ref{subsec:Plots}) for the Phoneme and SMS datasets (Supplementary Figs. S1-S6 contain plots of the remaining datasets). 
Pies on the main diagonal of Fig. \ref{fig:fig1}-A corresponds to a given performance measures while displaying the proportion of wins for each IAA technique under that metric. Pie charts in cells $(i,j)$, where $i\neq j$ can be used to compare metrics $i$ and $j$. The filled portion of the chart indicates agreement, while the empty section indicates disagreement.

For instance, when we examine Fig. \ref{fig:fig1}-A, it is evident that, in terms of Accuracy and AUC, both SMOTE and Upsample exhibit exceptional performances based on the respective pie on the main diagonal. However, F1, PR, and MCC indicate a preference for IPIP and ROSE. Interestingly, there is practically no agreement between F1 and AUC in terms of comparison, since the pie is mostly empty, while Accuracy and F1 measures seem quite similar. In Fig. \ref{fig:fig1}-B, we observe a similar situation. By examining the diagonal, it is evident that SMOTE outperforms the other methods in terms of Accuracy, Kappa, F1-Score, AUC, and MCC. However, when comparing pairs of different metrics, particularly the AUC for the Precision-Recall curve, it shows a notable lack of agreement with the others. This observation also applies to Balanced Accuracy. In addition, the plot shows SMOTE as the predominant method only in terms of Accuracy, Kappa, F1-Score, and MCC, but not for the AUC of the Precision-Recall curve. In addition, the plot shows SMOTE as the predominant method only in terms of Accuracy, Kappa, F1-Score, and MCC, but not for the AUC of the Precision-Recall curve. Therefore, the use of SMOTE does not always result in the best performance for the AUC of the Precision-Recall curve.

From the concordance plot, we can define a measure of discordance between two measures as the empty portion of a pie. In our experiments, we obtain a global discordance ratio across all measures of $0.66$, $99\%$ CI [$0.52$, $0.79$] in panel A, and $0.52$, $99\%$ CI [$0.39$, $0.66$] in Fig. \ref{fig:fig1}-B. Both this numbers and the plots in Fig. \ref{fig:fig1} clearly reflect that we cannot rely on these metrics to select the best imbalance-aware ML algorithm for a given dataset.

\subsection{All basic performance metrics are biased by the proportion of the minority class}
\label{subsec:Q2}
Once we have seen that there is no general consensus among metrics, it would make sense to ask whether, the metrics are distorted up to the point to show such a strong disagreement. This might be important if we want to develop an unbiased way to select the best imbalance-aware algorithm for a given problem. If we can estimate the level of distortion, we might use such estimates in our favour. Our idea is to integrate all metrics we know in a single, minimally biased metric that we can use to appropriately measure performance of IAAs \citep{2}.

\begin{figure}[ht!]
    \centering
    \includegraphics[width=0.8\linewidth]{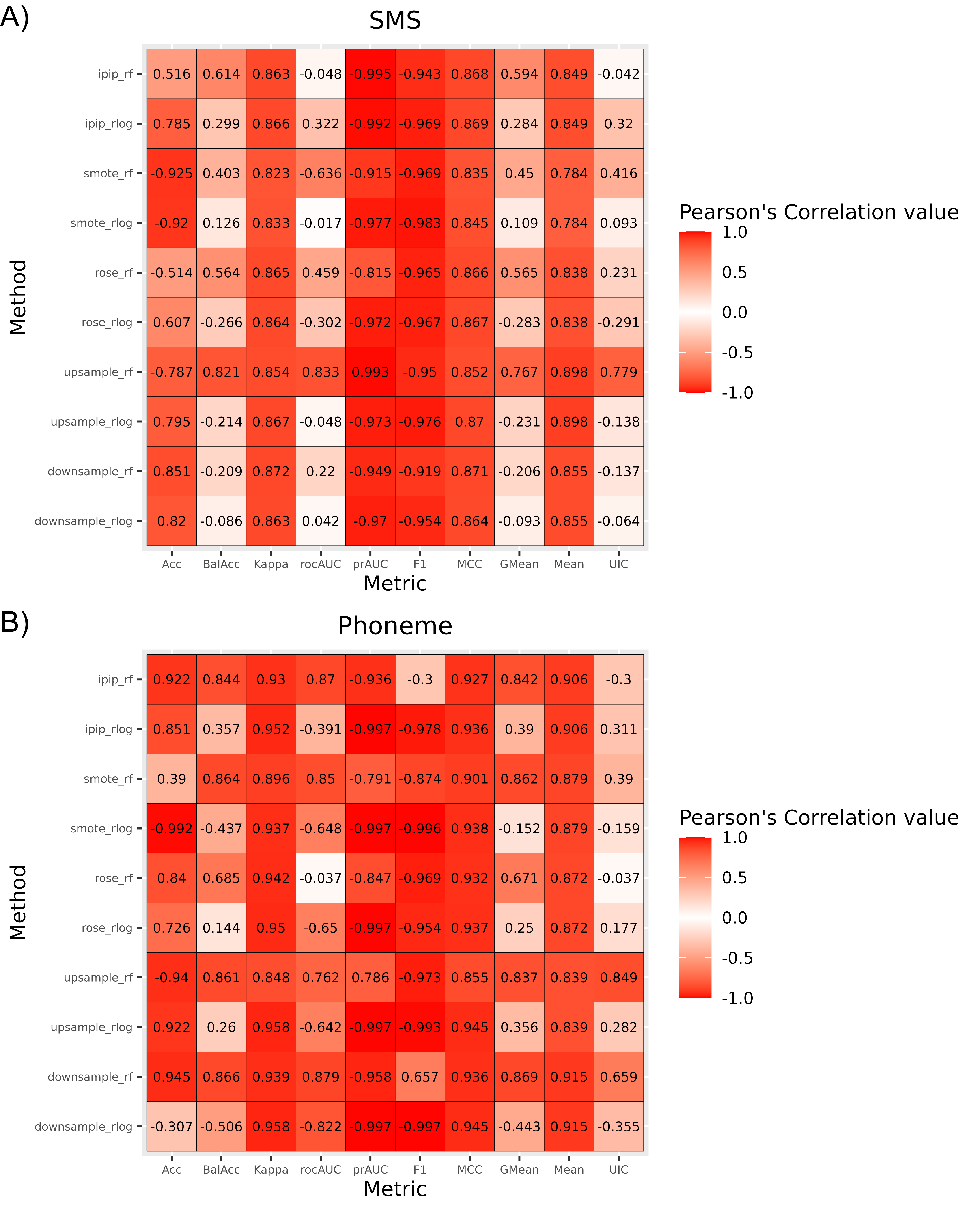}
    \caption{Correlation heatmaps beetwen metrics and $p_{min}$ in SMS (A) and Phenome (B) datasets.}
    \label{fig:fig2}
\end{figure}

To demonstrate the bias of common performance metrics due to the proportion of minority classes, we conducted a correlation analysis for each metric and $\mathbf{p_{min}}$. Fig. \ref{fig:fig2} A and B display the correlations between techniques (rows) and metrics (columns) in the form of a heatmap using a colour gradient from white (no correlation) to red (high absolute correlation, either positive or negative) for SMS and Phoneme datasets (Supplementary Figs S7-S11 contain plots of the remaining datasets). To the right of the heatmap, we included two new columns. One column shows the average of all the metrics, and the new metric, UIC, is presented in the other column.

In addition, Table \ref{table:distances} shows the Euclidean distance of absolute correlation values between each metric and dataset to the correlation vector \textbf{0}. This is further explained in Sec. \ref{subsec:Experiments2_Correlations}. Now, leaving the UIC corresponding column aside, and for the SMS dataset presented in Fig. \ref{fig:fig2}-A, the metric that shows the weakest correlation with $p_{min}$ is the area under the ROC curve (rocAUC), with a distance of 1.248 from vector \textbf{0}. Following this is the Geometric Mean (GMean) with a distance of 1.319. More specifically, the rocAUC demonstrates a stronger correlation within the Downsample model. The Random Forest algorithm achieves a Pearson's correlation value of 0.833, whereas in the Upsample model, it yields a correlation value of -0.636. Lastly, in the Rose model, it produces a correlation value of 0.459.

\begin{table}[!ht]
\resizebox{\textwidth}{!}
{
\begin{tabular}{|l|l|l|l|l|l|l|l|l|l|l|}
\hline          
                                           & \textbf{BalAcc} & \textbf{Kappa} & \textbf{Acc} & \textbf{F1} & \textbf{rocAUC} & \textbf{prAUC} & \textbf{MCC} & \textbf{GMean} & \textbf{Mean} & \textbf{UIC} \\ \hline
\multicolumn{1}{|l|}{\textbf{SMS}}         & 1.343           & 2.710          & 2.422        & 3.034       & 1.248           & 3.024          & 2.722        & 1.319          & 2.674         & 1.036        \\ \hline
\multicolumn{1}{|l|}{\textbf{Forest}}      & 2.014           & 2.897          & 2.381        & 2.809       & 2.262           & 2.145          & 2.895        & 2.025          & 2.768         & 1.243        \\ \hline
\multicolumn{1}{|l|}{\textbf{Phoneme}}     & 2.020           & 2.946          & 2.582        & 2.830       & 2.217           & 2.953          & 2.927        & 1.980          & 2.790         & 1.326        \\ \hline
\multicolumn{1}{|l|}{\textbf{Mammography}} & 1.308           & 2.594         & 2.654        & 3.062       & 1.171           & 2.911          & 2.621        & 1.317          & 2.320         & 1.033        \\ \hline
\multicolumn{1}{|l|}{\textbf{Satimage}}    & 1.632           & 3.045          & 2.720        & 2.783       & 1.499           & 2.996          & 3.025        & 1.532          & 2.873         & 1.304        \\ \hline
\multicolumn{1}{|l|}{\textbf{Diabetes}}    & 2.881           & 3.057          & 2.602        & 2.768       & 2.961           & 2.949          & 3.044        & 2.819          & 2.870         & 1.920        \\ \hline
\multicolumn{1}{|l|}{\textbf{Adult}}       & 2.347           & 2.796          & 2.952        & 3.106       & 2.153           & 3.031          & 2.913        & 2.356          & 2.690         & 1.816        \\ \hline
\end{tabular}
}
\caption{Distance matrix between correlation vectors for each metric and the vector $0$ in each dataset.}
\label{table:distances}
\end{table}

In the Phoneme dataset (refer to Fig. \ref{fig:fig2}-B), and again leaving aside UIC, the metric that is closest to 0, in terms of distance between the correlations and the vector \textbf{0} (shown in Fig. \ref{fig:fig2}-A), is the Geometric Mean (GMean), with a value of 1.980. This is followed by the Balanced Accuracy (BalAcc), with a value of 2.02. More specifically (See Fig. \ref{fig:fig2}-B), the GMean value correlates with $\mathbf{p_{min}}$ at 0.869 in Downsample with Random Forest, 0.862 in Smote with Random Forest, 0.842 in IPIP with Random Forest, and 0.837 in Upsample with Random Forest.  In general, based on Fig. \ref{fig:fig2}, it is not possible to conclude that there exists a metric that consistently has a bias towards 0 with respect to $\mathbf{p_{min}}$. This is due to the fact that the metric which is less biased in regards to the proportion of the minority class for the SMS dataset is different from the one that is less biased for the Phoneme dataset.

\subsection{A weighted integration of the metrics reduces bias}
\label{subsec:Q3}

\begin{figure}[ht!]
    \centering
    \includegraphics[width=1\linewidth]{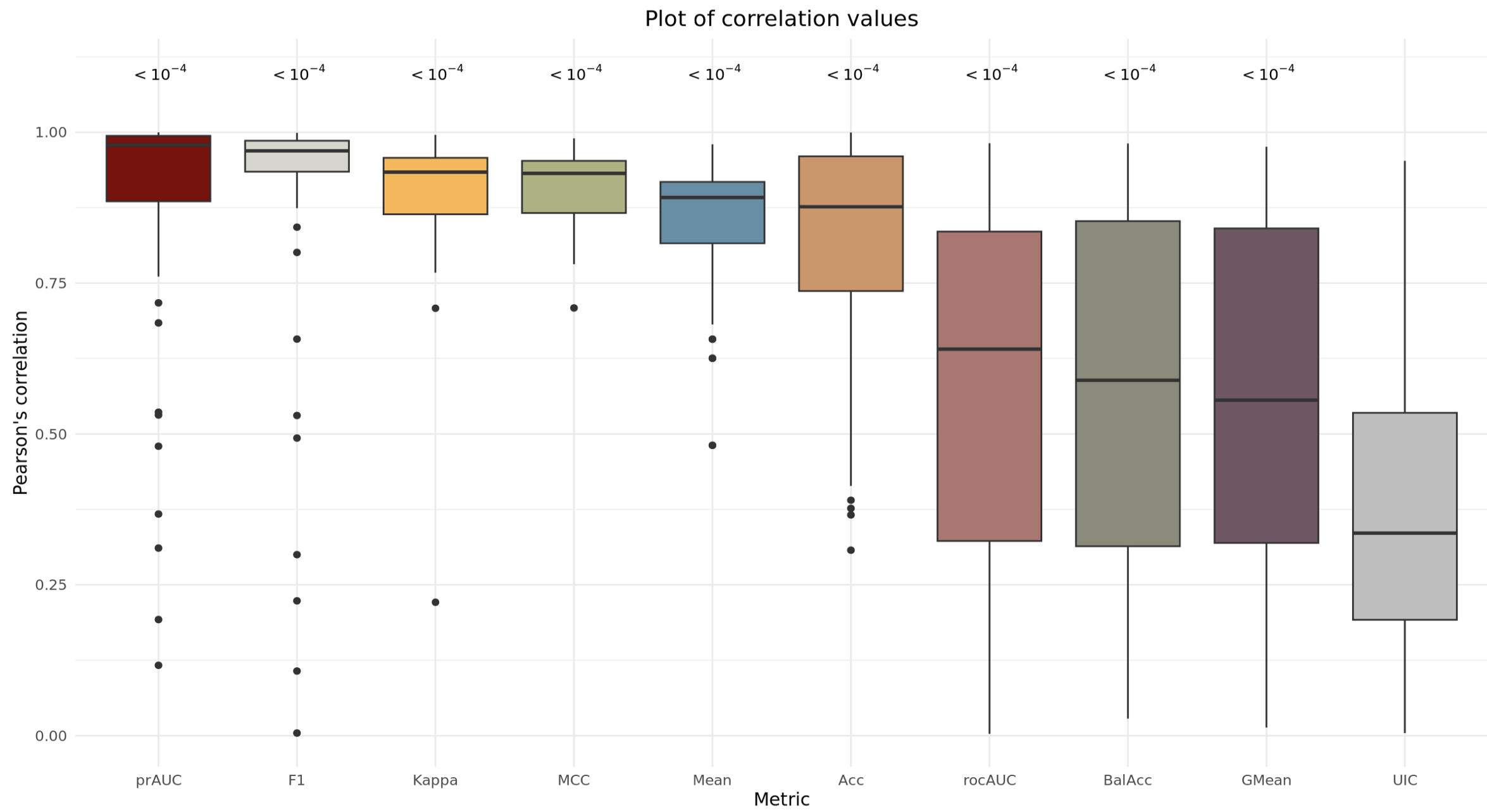}
    \caption{Boxplots of correlations of the different metrics across all experiments ordered by median from highest to lowest. Above each boxplot is the p-value of the Wilcoxon's Test vs UIC correlation values.}
    \label{fig:fig3}
\end{figure}

As introduced in Sec. \ref{subsec:UIC}, the correlation between the performance metrics $\mathbf{m_{dk}}$ and the minority proportion vector $\mathbf{p_{min}}$ is proposed as an estimate of the bias towards minority proportions. To examine the effect regardless of direction, Fig. \ref{fig:fig3} depicts a boxplot illustrating the distribution of absolute correlation values for each metric, including UIC and $\mathbf{p_{min}}$ obtained from all experiments.  For every metric, we present the p-value resulting from the Wilcoxon signed-rank test performed between all of the metrics and UIC, with Bonferroni correction applied. It is evident from the p-values (all $p<0.0001$) that there exist statistically significant differences, implying that UIC is less susceptible to bias towards $\mathbf{p_{min}}$.

\subsection{Results of UIC metric}
\label{subsec:Q4}

After determining that our experiments imply UIC values are less biased to $p_{min}$, we aimed to evaluate the performance of all IAAs, including IPIP, using UIC. Table \ref{table:2} illustrates the UIC for all five algorithms and all seven datasets, with logistic regression and Random Forest as base learners. The table does not demonstrate any conclusive outcomes regarding the best algorithm across all datasets. IPIP achieves the highest UIC value in three datasets. ROSE, Downsample, SmoteBoost and UnderBagging achieve the highest UIC value in a single dataset each. For the Diabetes dataset, which has one of the lowest number of variables in our experiments and the lowest number of instances, IPIP with RLOG achieves the highest UIC value. However, for datasets Forest and Satimage, which have the highest number of attributes (See Sec. \ref{subsec:Datasets}), IPIP with Random Forest obtains higher UIC values.

Interestingly, we detected a positive strong association between the number of attributes in the dataset and the UIC values achieved by IPIP when using random forest as base learner (correlation coefficient 0.64). This suggests that we should use random forest as IPIP's base learner on datasets with a large number of attributes. As one may had expected, this very same association is negative when IPIP is used with logistic regression (correlation coefficient -0.5). Therefore, logistic regression is still a good option as base learner in problems with a small number of attributes. IPIP is faster with this algorithm.

\begin{table}[ht!]
\resizebox{\textwidth}{!}
{
\begin{tabular}{|l|l|l|l|l|l|l|l|}
\hline
\multicolumn{1}{|c|}{\textbf{}}                  & \textbf{SMS}  & \textbf{Mammography} & \textbf{Forest} & \textbf{Satimage} & \textbf{Adult} & \textbf{Phoneme} & \textbf{Diabetes} \\ \hline
\multicolumn{1}{|l|}{\textbf{IPIP\_RF}}         & 0.91          & 0.00                 & \textbf{2.69}   & \textbf{0.73}     & 0.01           & 0.12             & 0.00              \\ \hline
\multicolumn{1}{|l|}{\textbf{IPIP\_RLOG}}       & 0.37          & 1.69                 & 0.01            & 0.13              & 0.00           & 0.10             & \textbf{0.78}     \\ \hline
\multicolumn{1}{|l|}{\textbf{SMOTE\_RF}}        & 0.03          & 0.31                 & 1.85            & 0.07              & 0.09           & 0.03             & 0.00              \\ \hline
\multicolumn{1}{|l|}{\textbf{SMOTE\_RLOG}}      & 2.25          & 0.69                 & 0.00            & 0.51              & 0.39  & 0.45             & 0.00              \\ \hline
\multicolumn{1}{|l|}{\textbf{ROSE\_RF}}         & 0.01          & 1.57                 & 0.02            & 0.52              & 0.11           & 0.86    & 0.00              \\ \hline
\multicolumn{1}{|l|}{\textbf{ROSE\_RLOG}}       & 0.47          & \textbf{1.85}                 & 1.48            & 0.00              & 0.00           & 0.65             & 0.03              \\ \hline
\multicolumn{1}{|l|}{\textbf{UPSAMPLE\_RF}}     & 0.00          & 0.00                 & 0.00            & 0.00              & 0.00           & 0.00             & 0.00              \\ \hline
\multicolumn{1}{|l|}{\textbf{UPSAMPLE\_RLOG}}   & 1.51          & 0.33                 & 0.04            & 0.23              & 0.01           & 0.21             & 0.00              \\ \hline
\multicolumn{1}{|l|}{\textbf{DOWNSAMPLE\_RF}}   & 1.02          & 0.63                 & 0.02            & 0.01              & 0.00           & 0.00             & 0.00              \\ \hline
\multicolumn{1}{|l|}{\textbf{DOWNSAMPLE\_RLOG}} & \textbf{2.42} & 0.93        & 0.29            & 0.00              & 0.00           & 0.10             & 0.00              \\ \hline
\multicolumn{1}{|l|}{\textbf{SMOTEBOOST\_RF}} & 0.00 & 0.00 & 0.01 & 0.00 & 0.00 & 0.00 & 0.02\\ \hline
\multicolumn{1}{|l|}{\textbf{SMOTEBOOST\_RLOG}} & 0.93 & 0.51 & 0.13 & 0.01 & \textbf{0.51} & 0.00 & 0.18\\ \hline
\multicolumn{1}{|l|}{\textbf{UNDERBAGGING\_RF}} & 0.37 & 0.68 & 0.01 & 0.00 & 0.23 & 0.02 & 0.54\\ \hline
\multicolumn{1}{|l|}{\textbf{UNDERBAGGING\_RLOG}} & 1.65 & 0.88 & 0.97 & 0.25 & 0.00 & \textbf{0.95} & 0.00\\ \hline

\end{tabular}
}
\caption{UIC values of the performance of the different models on the seven datasets.}
\label{table:2}
\end{table}

\subsubsection{Comparative of execution times for used ML algorithms}
\label{subsubsec:Complexity}

Table \ref{table:5} presents the time taken by every IAA method on each data set. It takes into account the average time spent implementing each technique, using the Logistic Regression and Random Forest ML models, in a five cross-validation procedure. IPIP is, on average, 16.25 times slower than the quickest alternative, Downsample. Note that SmoteBoost is the algorithm that requires the most time to complete on SMS, Mammography, Forest, and Satimage datasets, whereas IPIP is the longest-running algorithm on Adult, Phoneme, and Diabetes datasets. In contrast, Downsample, which trains a model on a smaller dataset, is the most straightforward method and takes the least time on all datasets.

\begin{table}[ht!]
\resizebox{\textwidth}{!}
{
\begin{tabular}{|l|l|l|l|l|l|l|l|l|}
\hline
                            & \textbf{SMS} & \textbf{Mammography} & \textbf{Forest} & \textbf{Satimage} & \textbf{Adult} & \textbf{Phoneme} & \textbf{Diabetes} & \textbf{Mean} \\ \hline
\textbf{DOWNSAMPLE\_RLOG}   & 1.11         & 0.50                 & 3.66            & 1.23              & 71.81          & 0.79             & 0.71              & \textbf{11.40}         \\ \hline
\textbf{ROSE\_RLOG}         & 32.36        & 1.73                 & 24.76           & 3.43              & 141.23         & 1.02             & 0.84              & \textbf{29.34}         \\ \hline
\textbf{SMOTE\_RLOG}        & 3.82         & 0.86                 & 15.08           & 2.25              & 229.61         & 1.67             & 1.32              & \textbf{36.37}         \\ \hline
\textbf{UPSAMPLE\_RLOG}     & 68.95        & 3.23                 & 45.20           & 5.97              & 192.95         & 1.25             & 0.85              & \textbf{45.49}         \\ \hline
\textbf{UNDERBAGGING\_RLOG} & 7.26         & 3.63                 & 26.70           & 6.47              & 321.70         & 4.63             & 5.39              & \textbf{53.68}         \\ \hline
\textbf{SMOTEBOOST\_RLOG}   & 167.87       & 10.78                & 203.37          & 21.12             & 94.02          & 8.70             & 4.47              & \textbf{72.90}         \\ \hline
\textbf{DOWNSAMPLE\_RF}     & 45.01        & 17.56                & 67.45           & 22.04             & 436.57         & 48.40            & 17.03             & \textbf{93.44}         \\ \hline
\textbf{IPIP\_RLOG}         & 88.67        & 22.15                & 79.35           & 30.55             & 1312.81        & 17.98            & 16.89             & \textbf{224.06}        \\ \hline
\textbf{SMOTE\_RF}          & 97.72        & 25.19                & 180.59          & 220.47            & 2152.45        & 93.10            & 24.45             & \textbf{399.14}        \\ \hline
\textbf{ROSE\_RF}           & 1166.42      & 116.60               & 466.19          & 66.48             & 1089.99        & 88.86            & 19.77             & \textbf{430.62}        \\ \hline
\textbf{UNDERBAGGING\_RF}   & 292.65       & 116.69               & 559.34          & 153.19            & 2367.74        & 313.43           & 113.09            & \textbf{559.45}        \\ \hline
\textbf{UPSAMPLE\_RF}       & 1801.88      & 108.32               & 497.46          & 78.40             & 1787.54        & 90.64            & 20.88             & \textbf{626.45}        \\ \hline
\textbf{IPIP\_RF}           & 822.10       & 397.58               & 1151.21         & 419.51            & 6434.39        & 794.04           & 338.84            & \textbf{1479.68}       \\ \hline
\textbf{SMOTEBOOST\_RF}     & 5396.05      & 565.39               & 2935.45         & 508.42            & 5652.97        & 602.18           & 143.41            & \textbf{2257.70}       \\ \hline
\end{tabular}
}
\caption{Elapsed mean time in second taken to train a 5 fold cross validation of each model ordered by column Mean, which is the mean time taken to train each model along all datasets.}
\label{table:5}
\end{table}

\section{Conclusions and future work}
\label{sec:Conclusions}


In this paper, we propose a new approach to tackle imbalanced datasets. Firstly, empirical investigations were conducted, revealing that most commonly used performance metrics exhibit inadequate concordance and an exaggerated inclination towards the minority class proportion. To address these issues, we introduce UIC, a new metric to measure the performance of Machine Learning algorithms on imbalanced data. 

The UIC measure, which encompasses various metrics analysed in this paper, exhibits a statistically significant lower correlation ($p < 10^{-4}$) with the proportions of minority classes. UIC is a suitable alternative strategy for the selection of the ideal model on a given data set. This innovative metric is especially advantageous in a production environment where changes in class proportions over time can adversely affect model performance. As UIC has a lower correlation with the imbalance ratio, it can assist in identifying a more effective ML model in cases where changes in class proportions degrade the performance of the model.

Secondly, we have also proposed IPIP, a novel IAA ensemble-based technique. IPIP involves obtaining resamples of the original dataset that are balanced in such a way that all instances of the minority class are represented in at least one of the samples. Another significant advantage of IPIP is that it does not generate synthetic instances. The algorithm has shown promising results as a viable alternative with higher UIC values in three out of seven datasets.  It is noteworthy that IPIP has the highest UIC using logistic regression as the base learner when working on the Diabetes Pima dataset, despite having the smallest number of instances. To ensure generalisation, it is advised to retain all instances across resamples, especially for smaller datasets where each instance holds significant importance. When IPIP is used in combination with Random Forest, the resulting values are notably better than those obtained by other state-of-the-art models, especially when the dataset contains a large number of attributes. Regarding IPIP execution time, it is considered efficient when compared to other imbalance-based methods. To calculate the unbiased UIC evaluation metric, a high computation time is required as experiments are performed on multiple auxiliary datasets to aggregate the basic performance metrics.

Among feature works, we plan to develop a new version of IPIP. This version will train different models simultaneously for each balanced resampled subset and then select the model with the best performance in each subset, instead of training one ML model in each subset. In addition, other ensemble approaches for models will be taken into account. Finally, the FILM\footnote{\url{https://github.com/antoniogt/FILM}} (Framework for Imbalanced Learning Machines) R package has been developed. This package provides code to build IPIP models, obtain the UIC metric, generate plots to compare metrics, and integrate all components. 

\backmatter

\bmhead{Acknowledgements}

This work was partially funded by Spanish Ministry of Science, Innovation and Universities (MCIU), the Spanish Agency for Research (AEI) and by the European Fund for Regional Development (FEDER) through CALM project (Ref: PID2022-136306OB-I00) and by the Science and Technology Agency, Séneca Foundation, Comunidad Autónoma Región de Murcia through the grant 21591/FPI/21. 

\section*{Statements and Declarations}

\begin{itemize}
\item Funding: This work was partially funded by Spanish Ministry of Science, Innovation and Universities (MCIU), the Spanish Agency for Research (AEI) and by the European Fund for Regional Development (FEDER) through CALM project (Ref: PID2022-136306OB-I00) and by the Science and Technology Agency, Séneca Foundation, Comunidad Autónoma Región de Murcia through the grant 21591/FPI/21.
\item Conflict of interest/Competing interests): Antonio Guillen reports financial support was provided by Fundación Séneca. Jose Palma reports
financial support was provided by Spanish Ministry of Science, Innovation and Universities. If there are other authors, they declare that they have no known competing financial interests or personal relationships that could have appeared to influence the work reported in this paper.
\item Ethics approval and consent to participate: All authors confirm their participation in the article.
\item Consent for publication: Not applicable
\item Data availability: All datasets are accessible from external sources except the SMS dataset which was retrieved from the Servicio Murciano de Salud (SMS). All SMS data produced in the present study are available upon reasonable request to the authors and the approval by Servicio Murciano de Salud (SMS).
\item Materials availability: Not applicable
\item Code availability: The code required to reproduce the FILM framework and results are available at: \url{https://github.com/antoniogt/FILM}.
\item Author contribution: A.G-T performed the formal analysis and all the experiments, wrote the main manuscript text and developed the R package. A.G-T., F.G. and M.C. developed the IPIP algorithm. A.G.-T., J.B. and J.P. developed the UIC metric. J.A.P. developed visualisation tools for the whole project. J.B. and J.P. supervised and directed the whole project. All authors participated in the paper writing up and all critically reviewed the manuscript.
\end{itemize}

\bigskip


\bibliography{sn-article}

\pagebreak
\includepdf[pages=-]{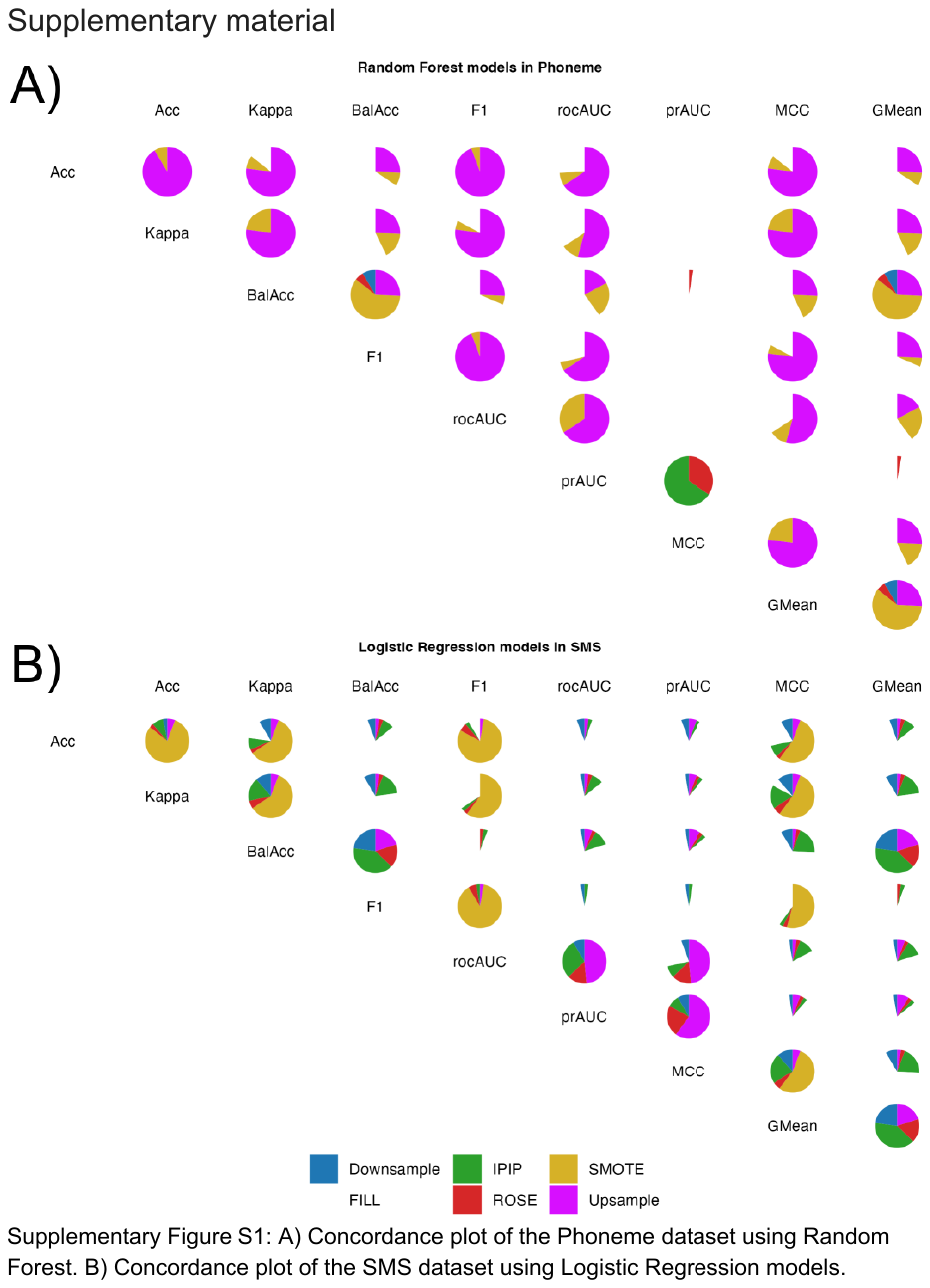}

\end{document}